\pdfoutput=1

\documentclass[10pt,twocolumn,letterpaper]{article}

\usepackage[pagenumbers]{iccv} 

\newcommand{\TODO}[1]{\PackageError{paper}{TODO has been disabled!}{dont do that}}

\definecolor{iccvblue}{rgb}{0.21,0.49,0.74}
\usepackage[pagebackref,breaklinks,colorlinks,allcolors=iccvblue]{hyperref}

\def\paperID{10481} 
\def\confName{ICCV}
\def\confYear{2025}

\title{Digitally Prototype Your Eye Tracker: Simulating Hardware Performance using 3D Synthetic Data}

\author{Esther Y. H. Lin\\
{\tt\small lin@cs.toronto.edu}
\and
Yimin Ding\\
{\tt\small yiminding@meta.com}
\and
Jogendra Kundu \\
{\tt\small jogendrak@meta.com}
\and
Yatong An\\
{\tt\small yatong@meta.com}
\and
Mohamed T. El-Haddad\\
{\tt\small mthaddad@meta.com}
\and
Alexander Fix \\
{\tt\small alexander.fix@meta.com}
}

\begin{document}
\maketitle
\begin{abstract}
Eye tracking (ET) is a key enabler for Augmented and Virtual Reality (AR/VR). 
Prototyping new ET hardware requires assessing the impact of hardware choices on eye tracking performance.
This task is compounded by the high cost of obtaining data from sufficiently many variations of real hardware, especially for machine learning, which requires large training datasets.
We propose a method for end-to-end evaluation of how hardware changes impact machine learning-based ET performance using only synthetic data.
We utilize a dataset of real 3D eyes, reconstructed from light dome data using neural radiance fields (NeRF), to synthesize captured eyes from novel viewpoints and camera parameters. 
Using this framework, we demonstrate that we can predict the relative performance across various hardware configurations, accounting for variations in sensor noise, illumination brightness, and optical blur.
We also compare our simulator with the publicly available eye tracking dataset from the Project Aria glasses, demonstrating a strong correlation with real-world performance.
Finally, we present a first-of-its-kind analysis in which we vary ET camera positions, evaluating ET performance ranging from on-axis direct views of the eye to peripheral views on the frame.
Such an analysis would have previously required manufacturing physical devices to capture evaluation data.
In short, our method enables faster prototyping of ET hardware.
\end{abstract}    
\section{Introduction}
\label{sec:intro}

Eye tracking (ET) is a cornerstone of AR/VR and enables applications like foveated rendering and gaze-based interactions~\cite{jin2024eye,patney2016towards,piumsomboon2017exploring}.
Building ET cameras for AR/VR devices is a challenging dance: camera specifications and placement directly impact performance~\cite{holmqvist2011eye}, while form-factor and near-eye design constraints impose tradeoffs.
In this work, we leverage data-driven ML ET pipelines~\cite{engel2023project} to accelerate this challenging design process, due to their proven scalability and robustness.
Our goal is to enable rapid prototyping of ET camera designs and placements, and evaluate them using ML-based pipelines.
However, this goal faces significant obstacles.
First, the time and monetary cost associated with acquiring the training data needed to evaluate each early design is substantial.
Second, ET ML models are inherently tied to the specific hardware configuration on which they were trained~\cite{ghosh2023automatic}, and there is currently no reliable method to extrapolate their performance to alternative hardware setups.
Overall, these challenges necessitate a costly procedure of collecting distinct real datasets for each hardware configuration under consideration.

Synthetic data generation aims to alleviate the high cost of acquiring real data.
We exploit this to reduce the high cost of iterating on ET hardware design. 
For ET, previous synthetic data leveraged meshes to reconstruct the eye and surrounding facial regions~\cite{wood2016_etra,porta2019u2eyes,nair2020rit,bermano2015detailed}.
While mesh-based methods can support arbitrary camera views, they struggle to capture local high-frequency details, such as eyelashes, which are critical in ET.
More recently, EyeNeRF~\cite{li2022eyenerf} leveraged 3D implicit models to produce visually-pleasing and detailed reconstructions of the eye, but did not address synthetic data generation or improved ET as a use case.

\begin{figure*}[!ht]
    \centering
    \includegraphics[width=0.95\textwidth]{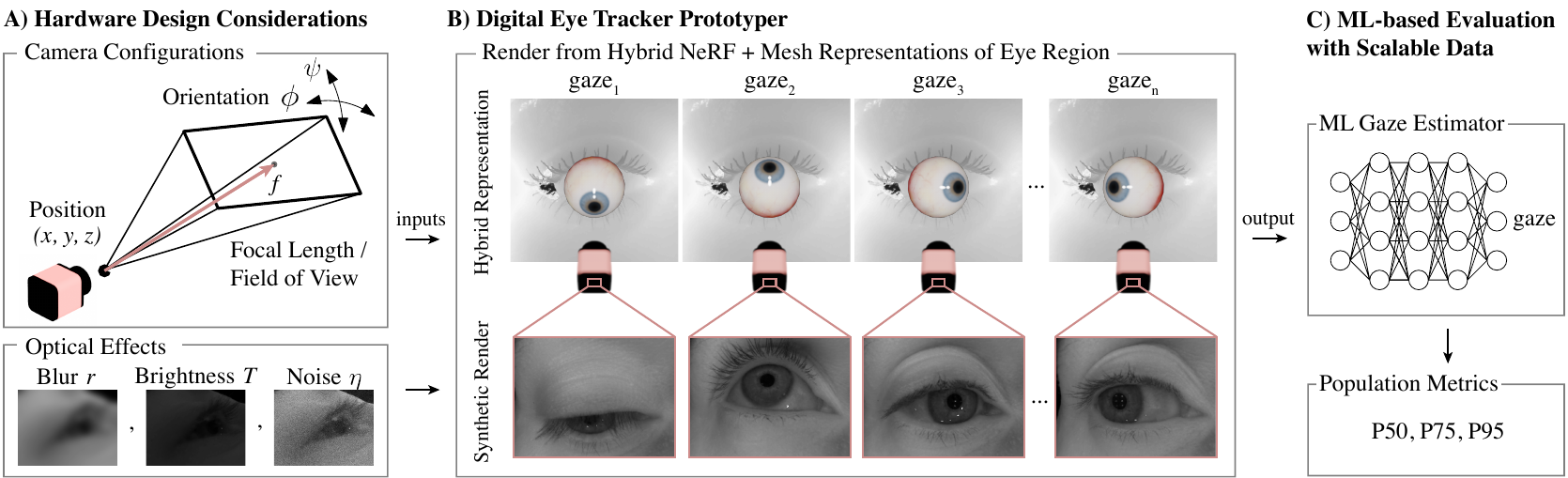}
     \caption{
    \textbf{Overview.} 
    The \textit{Digital Eye Tracker Prototyper} receives (A) hardware design considerations—both camera configurations and optical effects—as input.
    It employs a collection of (B) 3D hybrid representations (NeRFs and meshes) of the eye and periocular region to render images for each gaze. 
    There are 114 gaze representations per identity across 195 identities in the digital eye tracker prototyper. 
    The renders are fed to (C) an evaluator that trains a deep gaze estimator and evaluates performance for the given hardware design considerations.}
    \label{fig:system-diagram}
\end{figure*}

Our approach adapts these 3D implicit methods to generate ET-specific synthetic data.
It features a two-component simulator: (1) a mesh-NeRF hybrid representation for eye modeling,
tailored to ET applications with considerations such as illumination wavelength and high resolution on the eye; and (2) an optical simulator that incorporates effects such as blur, brightness, and signal-to-noise ratio (SNR) from the ET camera.
We demonstrate that our simulator predicts relative performance metrics for ET hardware architecture choices when its renders are used as training data for ET ML models.
We emphasize that this work does not seek to innovate on ML models for eye tracking. 
Instead, we use a fixed state-of-the-art ML model from the open-source Project Aria~\cite{engel2023project} and show that we can predict how the model will perform as we change camera parameters and optical properties of the underlying ET hardware.
Our findings show that our synthetically-trained models are highly correlated in performance with real-data-trained models, as a function of various camera properties.
This means, if a simulated hardware change degrades performance by X\%, we expect real hardware performance to shift by approximately the same amount.
This is useful for quick, scalable comparisons between prototype ET hardware designs.

More broadly, evaluating computer vision hardware designs is a nuanced problem because designs are indirectly influenced by downstream applications. 
Optical metrics such as the modulation transfer function and SNR are traditionally used to evaluate camera designs because they offer explicit design rules.
However, when deep, black-box ML models evaluate aggregate system performance, deciphering their contributions becomes challenging because the design signals are implicit.
We bridge this gap by developing an optical simulator to probe ML-based ET performance using various optical metrics as input.
 
In summary, our contributions include:
\begin{itemize}
    \item A light dome capture setup tailored for eye tracking that provides reconstructions of real identities and the ability to synthetize novel views of those identities for arbitrary camera configurations. 
    \item Demonstration through multiple comparisons between real- and synthetic-trained models that this simulator gives highly correlated relative signal on eye tracking hardware architecture choices.
    \item A first-of-its-kind evaluation of how camera position affects ET performance, enabled by our framework's ability to support novel view synthesis.
\end{itemize}

\section{Related Works}
\label{sec:related}

\textbf{Evaluating Eye Trackers} Eye tracker evaluations have primarily focused on gaze estimation algorithms~\cite{swirski2014rendering, kim2019nvgaze},
leaving hardware parameters like camera placement and optical effects largely unexplored.
To the best of our knowledge, we present the first framework for assessing hardware and optical considerations in eye trackers.

A starting assumption in hardware design is that images with higher optical quality result in better performance. 
But the key question is ``how much?'' 
Prior computer vision works show that better performance on optical metrics does not guarantee improved outcomes in tasks outside eye tracking, such as image matching~\cite{chen2025isotropic}, depth estimation~\cite{yang2023aberration}, automotive object detection~\cite{tseng2021differentiable}, or hyperspectral imaging~\cite{shi2024learned}, especially when the captured data is ultimately fed into an ML-based evaluation system, which can exhibit nonlinear responses to input quality. 
We bridge this gap by developing an optical simulator that probes ML-based eye tracker performance using synthetic data.

\noindent\textbf{End-to-End Gaze Estimation}
There are two overall approaches to gaze estimation.
Traditional 3D geometric gaze estimation approaches relied on detecting eye features and combining them with an eye geometry model (e.g., an elliptical model) to extract eye positions and estimate gaze~\cite{li2005starburst, kassner2014pupil, fuhl2016else, fuhl2015excuse}.
These approaches are difficult to extend to arbitrary camera positions and often lack robustness with occlusions, poor pupil visibility, or strong specular reflections from surrounding skin~\cite{javadi2015set,swirski2014rendering, kassner2014pupil, nair2020riteyes}.
Later ML-based  approaches, particularly those based on convolutional neural network (CNN) architectures, emerged as state-of-the-art for gaze estimation~\cite{kim2019nvgaze, wood2015rendering, wood20163d, wood2016learning, nair2020riteyes, zhang2015appearance}.
They demonstrate robust performance under challenging conditions, including extreme gaze angles, poor pupil visibility, and strong reflections~\cite{kim2019nvgaze, wood2016learning, nair2020riteyes}.
Notably, the open-source Project Aria ET model~\cite{engel2023project}, which employs a ResNet-18 backbone, has been validated on thousands of real-world users~\cite{pan2023ariadigitaltwinnew, engel2023project}.
We find the end-to-end gaze estimation approach most suitable for evaluating our hardware variation experiments, as its minimal input requirements allow for direct application across arbitrary camera positions.\par 

\noindent\textbf{Gaze Datasets} Training CNNs for gaze estimation requires a large corpus of annotated eye images~\cite{nair2020riteyes, garbin2019openeds}.
Publicly-available labeled datasets specific for AR/VR gaze estimation purposes include
OpenEDS, with 12,759 off-axis images of real eyes, captured using infrared (IR) eye cameras in a head-mounted display~\cite{garbin2019openeds}; and 
NVGaze with 2 million synthetic eye images with a near-eye camera configuration~\cite{kim2019nvgaze}.
However, there is no clear method to adapt these datasets to simulate ET cameras at arbitrary positions.

\noindent\textbf{Eye and Periocular Region Modeling for Synthetic Data Generation}
We look at methods which support synthesizing data for arbitrary cameras.  
We want methods based on real captures of human eyes to minimize the domain gap between synthetic and real datasets.

Previously, parametric models that use ellipses for the pupil~\cite{li2005starburst, kassner2014pupil, fuhl2016else, fuhl2015excuse, hoshino2020estimation}, or spheres for the eyeball~\cite{grand1957light}, 
have been desirable due to their explicitness.
However, they cannot model the surrounding periocular region or capture the full complexity of the eye~\cite{sagar1994virtual}.

Mesh-based representations, however, can model the anatomical attributes of the eye~\cite{berard2014high, berard2016lightweight, berard2019eyerig} 
and the periocular area~\cite{neog2016interactive, schwartz2020eyes}, as well as support rendering from arbitrary camera views.
Meshes of the eye and facial regions are readily available in the 3D artistry domain\footnote{\url{https://blendswap.com/blend/5877}}, where high-resolution scan assets can be used to generate data~\cite{nair2020riteyes}.
However, mesh-based representations struggle to model fine structures such as eyelashes~\cite{li2022eyenerf} and do not fully capture the complex reflectance of the eye and surrounding skin~\cite{nishino2006corneal, alzayer2024seeingworldeyes},
limiting their ability to generate high-quality synthetic images under controlled lighting conditions~\cite{li2022eyenerf}.
Artist-application-driven meshes may also lack anatomical accuracy.
For example, capturing eyelash details and simulating the skin's IR response require extra modeling~\cite{swirski2014rendering, nair2020riteyes, kim2019nvgaze} that may not fully reproduce the realism and diversity of actual features.

Implicit volumetric representations solve many of these challenges:
NeRFs~\cite{mildenhall2020nerf} showed that neural network-based volumetric reconstructions from multi-view images can achieve photorealistic novel view synthesis while retaining high-frequency details~\cite{tancik2020fourfeat}.
EyeNeRF~\cite{li2022eyenerf} combines an implicit deformable volumetric representation for face and eye interior with 
an explicit mesh representation for the eyeball surface, allowing it to capture complex reflectance and fine-scale geometry for animatable and relightable 3D avatars.
The present work builds upon ideas from EyeNeRF to create an eye-tracking-specific system by employing near-infrared (NIR) cameras, maximizing resolution on the eye, and capturing a dense sampling of gaze directions.

\noindent\textbf{Synthetic Data Domain Gap} 
Synthetic data typically has a domain gap with real data. 
For example, NVGaze~\cite{kim2019nvgaze} demonstrated that models trained solely on synthetic data do not perform equivalently when evaluated on real data.
This gap can be alleviated by augmenting the synthetic data with a few real-world images~\cite{Park2019ICCV} or by using domain adaptation methods to close the sim-to-real gap~\cite{nguyen2024deep}.
However, all these approaches require at least some real data.
In the case of early hardware prototyping, getting even one image from a new hardware configuration requires a complete hardware build.
Instead the present work focuses on what can be done with zero real data, to predict the relative performance resulting from hardware changes.

\section{Method}
\label{label:method}

To shortcut the process of predicting performance of prototype ET hardware,
we present a digital twin framework~\cite{phanden2020digitaltwin},
where we simulate a digital version (digital twin) of the desired hardware, generate a synthetic training set appropriate for training an ML ET model, and then estimate the performance of the hardware change from the performance of the synthetically-trained model.
Our synthetic data generator, based on data-driven 3D reconstructions of real eyes, enables the exploration of various parameters—including optical characteristics, camera placements, lighting configurations, and even glasses slippage.
The framework is illustrated in Fig.\ref{fig:system-diagram}.

We discuss the capture setup for 3D reconstructions of real eyes (Sec.~\ref{ssec:setup}), constructing hybrid representations for each gaze (Sec.~\ref{ssec:hybrid}), and the image formation model for incorporating optical effects (Sec.~\ref{ssec:optics}). Finally, we leverage our digital twins to evaluate ET performance of various hardware design choices by training and end-to-end ML models on each dataset (Sec.~\ref{ssec:ml-eval}).

\subsection{Capture System}
\label{ssec:setup}

In this section, we describe the choices made to build a light dome specifically targeted to 3D reconstruction for synthetic eye tracking data. 
At a high level, the system maximizes the number of pixels on the eye, uses the same near infrared (NIR) wavelengths as eye tracking hardware, and includes static gaze targets to get a repeatable large set of gaze directions for each person scanned.

\begin{figure}[ht]
    \centering
    \includegraphics[width=0.48\textwidth]{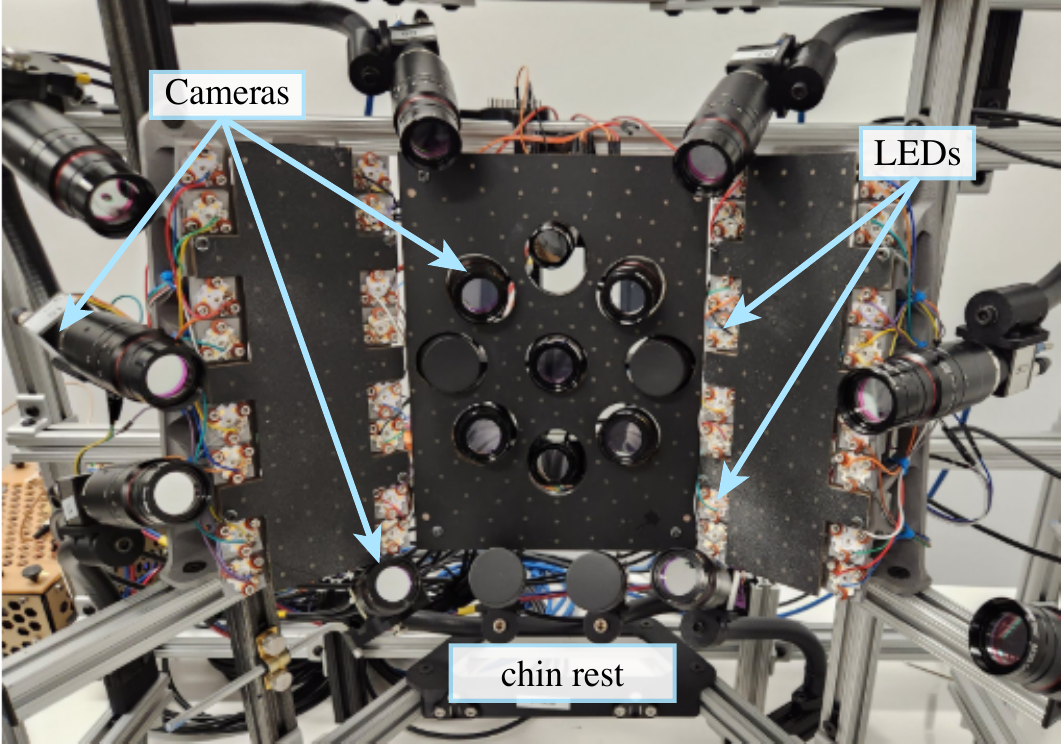}
    \caption{
    \textbf{Light dome capture setup } 
    with cameras, NIR LEDs and a chin rest rigidly mounted to a frame.
    Captures are used to construct 3D models of the eye and periocular region. }
    \label{fig:light-dome}
\end{figure}

The setup (Fig.~\ref{fig:light-dome}) is composed of 20 NIR cameras (model type: IDS UI-3370CP-NIR) and 16 point lights of NIR LEDs at 850 nm (OSRAM LZ1-10R702), held together rigidly in a desk-mounted frame.
A custom LED driving circuit was designed to maintain an irradiance \textless 9 W/m\textsuperscript{2 } at the nominal pupil position, well below the eye-safe limit of 25 W/m\textsuperscript{2} for chronic exposure set by the IEC62471 standard. 
A narrowband filter (Edmund Optics Inc, 66-235)  is applied to the cameras to capture only 850 nm light.
As noted in~\cite{kim2019nvgaze}, there is a domain gap between RGB images and NIR images used for eye tracking, so it is important to capture at the final wavelength used for dataset generation.
The setup additionally features a chin rest to minimize head movement, with participants aligned via translation stages to the center of field of view and with iris centered in the depth of field.

The number of cameras and placement were chosen to maximize the number of pixels on the eye and periocular region visible from in-headset eye tracking cameras, while fitting within the USB bandwidth and disk-write limits of a single collection PC.
Due to mechanical constraints of fitting the required cameras in the collection rig, only the participant's right eye (OD) is captured. 
The left eye (OS) is rendered by mirroring the 3D reconstruction.

The camera intrinsics and extrinsics are calibrated before each collection session using a robotic calibration target, and light positions are calibrated using a reflective sphere.
This calibration is used during the NeRF reconstruction instead of using COLMAP~\cite{schoenberger2016sfm, schoenberger2016mvs} for bundle-adjustment on the scene, as is often done as the first step of NeRF. 

To prompt the user to look at a diversity of gaze directions, we use 114 gaze targets on a flat board with cut-outs for the cameras.
These are arranged within a $\pm35^\circ$ pitch and yaw field of view. 
The user is prompted to look at each target sequentially via a visible-light LED at each target location. 
While the participant is gazing at a target, all cameras capture images of the eye from different angles. 
This approach differs from the camera-on-a-stick method employed by EyeNerf~\cite{li2022eyenerf} for gaze targets, providing a more controlled and repeatable environment for data collection.

\subsection{3D Reconstruction and Rendering}\label{ssec:hybrid}
From our captures, we develop a simulator that can render images of the eye from any viewpoint. 
This allows flexible and realistic simulation of various ET scenarios.

\noindent\textbf{Hybrid representation}
We construct a hybrid representation for each gaze target using our captures, similar to EyeNerf~\cite{li2022eyenerf}. 
This representation is termed ``hybrid" because it leverages both NeRFs and meshes to reconstruct the eye and its surrounding regions.  
We depart from the approach of EyeNeRF by performing 114 static reconstructions at each gaze direction for each identity, rather than learning one dynamic (animated) reconstruction per identity.
In early prototyping of our method, we found that the static reconstructions had higher image quality than dynamic models, and that 114 selectable gaze directions per identity was sufficient for generating synthetic training data.

For the mesh, we utilize a personalized 3D eyeball, obtained through a separate 3D scan of the eye
to account for refractive effects during intraocular ray tracing and to explicitly calculate specular reflections of light rays, i.e., the glints. To ensure precise ocular pose estimation, we employ a geometric ET pipeline that optimizes the 3D eye position by solving for the rigid transformation of the eyeball mesh through simultaneous refinement of pupil boundary correspondences and glint alignment across camera viewpoints.

\noindent\textbf{Rendering with varying camera configurations} 
To simulate novel hardware, we input camera viewpoints (specified in device coordinates) into the hybrid representation, which then renders images at the specified viewpoints.
The simulated camera positions are taken from real device calibrations, e.g., those from Aria glasses available at~\cite{projectaria_tools}.
Novel cameras are simulated by inputting the device parameters from optical/CAD design software.

Head mounted eye trackers are designed with a nominal ``fitment'' of the device on the user's head. 
When donning the eye tracker, users do not position the device precisely at the nominal position due to preference in ``device fit'',
and normal motions of the head and body cause the headset to ``slip'' relative to this initial position. 
We simulate fitment and slippage by introducing a transformation matrix between ideal device fit and simulated device fit,
with the amount of translation sampled from a distribution obtained by measuring displacements during real usage.

\subsection{Optical Effects Simulator}\label{ssec:optics}

After rendering images from our 3D reconstructions, we incorporate optical effects with an optical simulator. 
Traditionally, optical metrics like the modulation transfer function (MTF) and signal-to-noise ratio (SNR) provide explicit design rules for camera systems~\cite{kidger2001fundamental,sun2015lens,yang2017automated}. 
These factors manifest in three key eye tracker design parameters: blur, noise, and brightness. 
Due to form factor, cost, and power constraints, eye trackers use small, inexpensive cameras that often suffer from significant blur, high noise, and low brightness.
The optical simulator incorporates these optical effects into generated synthetic data. 

We model an output image from the optical simulator as:
\begin{equation}
    I(x, y) = \int_{0}^{T}  O(x,y) \circledast \text{PSF}(x, y)  \, dt + \eta,
    \label{eq:image_formation}
\end{equation}
where $O(x,y)$ is the linear scene irradiance rendered from our hybrid representations, $\text{PSF}(x, y)$ is the point spread function, $\circledast$ denotes convolution, $\eta$ represents noise, and $T$ is the exposure time. This image formation process is linear since we render outputs to a \texttt{float32} array instead of quantizing to 8-bit images.
This enables later linear rescaling for brightness modulation without loss of precision.

\begin{figure}[h]
    \centering
    \includegraphics[width=0.98\linewidth]{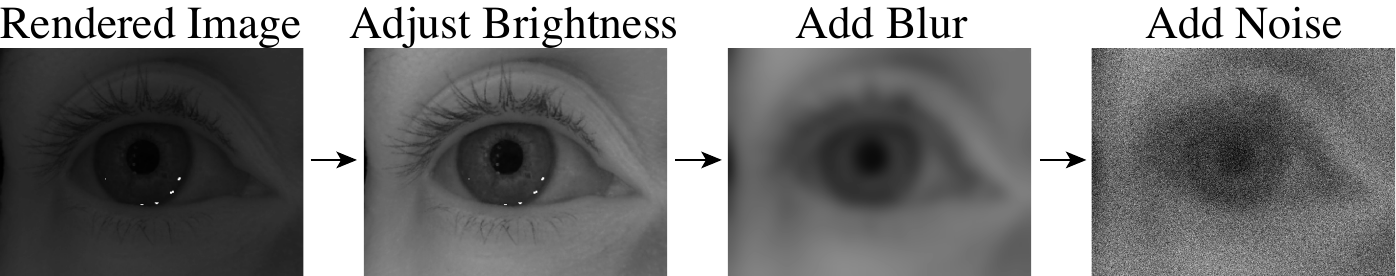}
    \caption{\textbf{Order of operations in optical simulator.} Input is the linear rendered image from the hybrid reconstruction. The output is an image with modified brightness, aperture blur, and noise.}
    \label{fig:optical-simulator}
\end{figure}

Next, we discuss the sequence of operations in our optical simulator, as depicted in Fig.~\ref{fig:optical-simulator}.

\noindent\textbf{Brightness} Before applying noise and blur, we adjust the brightness of the digital twin render by leveraging its linear \texttt{float32} format.
This enables a linear adjustment of \(O(x,y)\) via scalar multiplication.
In physical AR glasses, brightness is controlled by ambient illumination, ET LED intensities, and camera exposure.

\noindent\textbf{Blur} Next, we apply spatially-invariant aperture blur to the image, utilizing the Fourier domain and the convolution theorem for efficient computation. We perform blurring on the linear \texttt{float32} image—before incorporating sensor saturation modeling—thereby capturing the effect of blur from bright specularities frequently observed on the eye.

\noindent\textbf{Noise} Finally, we apply noise. We follow the physics-inspired noise model derived from Monakova et al~\cite{monakhova2022dancingstarsvideodenoising} and Hasinoff et al~\cite{hasinoff2010noise}, modeling noise $\eta$ as
\begin{equation}
    \eta = \eta_\text{shot} + \eta_\text{read} + \eta_\text{thermal} + \eta_\text{quantization} + \eta_\text{fixed pattern},
    \label{eq:noise_sources}
\end{equation}
where $\eta_\text{shot}$, $\eta_\text{read}$, $\eta_\text{thermal}$, $\eta_\text{quantization}$, and $\eta_\text{fixed pattern}$ approximate the contributions of shot noise, read noise, thermal noise, quantization noise, and fixed pattern noise, respectively. 
Shot and thermal noise can be modeled with a Poisson random variable, while read noise is modeled with a normal distribution~\cite{hasinoff2010noise}.

We experimentally verify the contributions of these noise sources with the camera sensors in Project Aria glasses. 
We find that for the Project Aria glasses, the contributions of $\eta_\text{thermal}$ and $\eta_\text{fixed pattern}$ are negligible. 
Additionally, by the central limit theorem, the Poisson distribution converges to a normal distribution for large event counts, which allows us to simplify our noise model to
\begin{equation}
\eta = \eta_\text{normal} + \eta_\text{quantization},
\end{equation}
where $\eta_\text{normal} = \eta_\text{read} + \eta_\text{shot}$. We apply $\eta_{\text{normal}}$, modeled as a Gaussian random variable. As the final step before saving the image, we add $\eta_{\text{quantization}}$ using an 8-bit quantization scheme consistent with the Project Aria camera sensors.

\subsection{Machine Learning Performance Evaluation}
\label{ssec:ml-eval}

\noindent\textbf{Project Aria gaze estimation model}
For eye tracker evaluation, we adopt a CNN-based gaze estimation approach, as these are agnostic to camera view, ensuring that the synthetic data generated from our method can be evaluated across a range of setups. 
We use the state-of-the-art open-source Project Aria ET model~\cite{engel2023project} to evaluate our hardware designs. 
The model features a ResNet-18-based CNN backbone, accepts input images of size $240 \times 320$ pixels for both the left and right eyes, and outputs an estimated 3D gaze vector.

\noindent\textbf{Training details} 
We train independent Project Aria gaze estimation networks on each of our datasets. 
We use the default hyperparameters provided by Project Aria: 50 epochs, $\text{batch size} = 64$, $\text{learning rate} = 5\times10^{-4}$, and Adam optimizer. 
Each model is trained on a single NVIDIA A100. 

\section{Experiments}
\label{label:experiments}

In this section, we present results demonstrating that synthetically-trained models are predictive of relative differences in model performance with respect to changing camera parameters:
when synthetic data shows performance degradation of X\% as a camera parameter is changed, then real-data trained models show a similar degradation. 
In all experiments, the only variable that is changed is the training set and test set used for evaluation. 
All other hyperparameters of the model are held constant. 
By isolating the effect of the dataset on the model performance, a high correlation indicates that synthetically-trained models are predictive of relative difference in real-trained model performance.

Note that we still see a domain gap between real and synthetic data: models trained on synthetic data do not have good performance when evaluated on real data, and vice versa.
Additionally, there is an overall scale difference in the performance metrics between synthetically-  and real-trained models, which we attribute to different ``difficulty'' in the overall prediction tasks.

\subsection{Comparing Real- and Synthetic-trained Model Perfomance}
\label{ssec:compare-to-real}

\begin{figure}[h]
    \centering
    \includegraphics[width=0.96\linewidth]{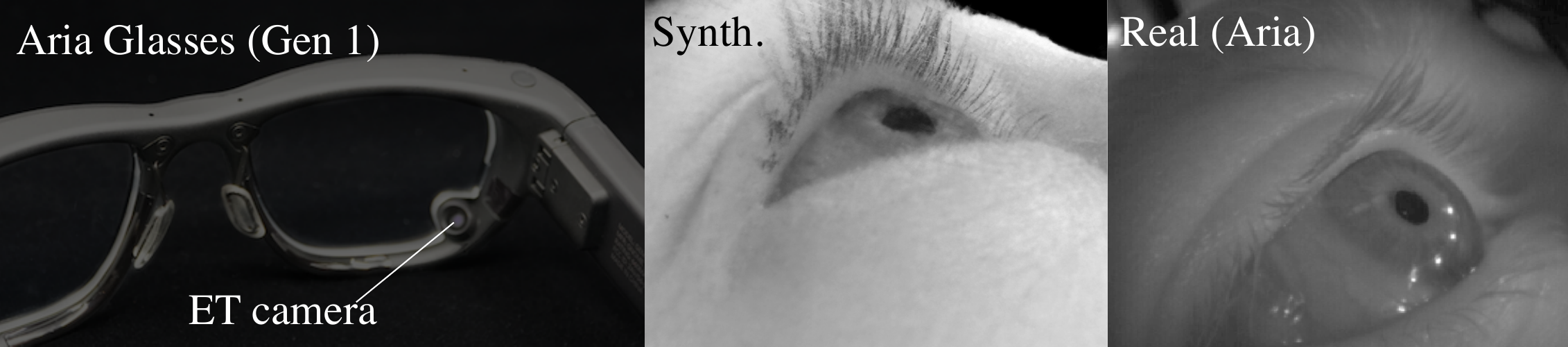}
    \caption{\textbf{Synthetic and Real Images.} We compare synthetic (Synth.) images simulated using the configurations of the ET camera on Aria glasses with real ET images captured by the glasses. The synthetic images are able to capture details such as eyelashes and skin texture.}
    
    \label{fig:synth-v-real}
\end{figure}

\subsubsection{Comparison metrics}
To compare the results from trained models, we compute the average and standard deviations over 3 trials for all experiments, with each trial trained on a different randomized training and test set split. 
We report the 50th, 75th, and 95th percentiles of gaze error respectively for all results.
We particularly emphasize the 95th percentile results, because users will encounter many such errors during a session of typical eye tracking applications (such as foveated rendering and gaze-based interactions~\cite{jin2024eye,patney2016towards,piumsomboon2017exploring}), and these dominate the quality of user experience.

\subsubsection{Datasets}
For a summary of dataset sizes and number of identities, please refer to Tab. \ref{tab:dataset_sizes}.

\paragraph{Real-life dataset} We leverage the open-source dataset from Project Aria, which includes ET camera images from over ~2100 identities with ~6.11 million frames.
Each frame is labelled with a 3D gaze vector.
We split the Project Aria dataset into three non-overlapping datasets (Big Aria, Small Aria, Aria Test) to control the effect of dataset size and number of identities on training performance.
Because we are limited in the number of real identities we were able to scan in our light-dome, the Small Aria dataset was created to have an equal number of identities for training between the synthetic and real datasets.
Then, the Big Aria dataset contains 10 times the number of identities to compare the effect of having much greater identity diversity. 
Finally, a separate Aria test set is created for evaluation, with a ratio of 1:4 compared to the Big Aria set. 
This split allows for a comprehensive evaluation of model performance on both small and large datasets.

\paragraph{Synthetic Dataset}
Our synthetic dataset comes from scans of 195 identities, each with hybrid reconstructions of 114 gaze targets. 
We employ a 4:1 training and testing dataset split on the identities.
Test sets are composed of identities unseen during training.
In order to create a dataset with a comparable number of images to the with the real-life Small Aria dataset, we render 24 uniformly-sampled random slippage variations $(\Delta x, \Delta y, \Delta z)$ in 3D per gaze target, 
giving a total of 2,736 images per identity for the synthetic data.
This achieves a similar total number of identities and total images for the Synth Train and Small Aria datasets (Tab.~\ref{tab:dataset_sizes}).
Our slippage ranges of $\Delta x, \Delta y \in [-3, 3] \text{mm}$ and $\Delta z \in [-6, 6] \text{mm}$ cover 2 standard deviations of slippages extracted from Aria accelerometer data.
Finally, we take camera calibrations provided by Project Aria~\cite{projectaria_tools} and input them into our framework to generate images at Aria ET camera viewpoints in device coordinates.
A visualization of synthetic and real images is given in Fig.~\ref{fig:synth-v-real}.

\begin{table}[t]
  \centering
  \resizebox{0.38\textwidth}{!}{
   \begin{tabular}{l|c|c}
         Dataset & Identities & Number of Images (M) \\
         \hline\hline
         Big Aria         & 1550 & 4.11 \\
         Small Aria       & 155  & 0.46 \\
         Aria Test        & 388  & 1.11 \\
         Synth Train & 155  & 0.42 \\
         Synth Test  & 40   & 0.11 \\
  \end{tabular}
  }
  \caption{\textbf{Dataset Sizes.} Number of identities and images (M = million) per dataset. Each identity in all datasets (both synthetic and Aria) has an average of 2727 frames.
  }
  \label{tab:dataset_sizes}
\end{table}

\subsubsection{Baseline comparison}
We first evaluate the the performance of models trained on synthetic data versus models trained real-life Aria data with no additional optical effects from Sec.~\ref{ssec:optics} applied.
We train a model on the Big Aria, Small Aria, and Synth datasets, and test on both the Aria and Synth test sets.
Results are reported in Tab. \ref{tab:base_case_results}. 

\begin{table}[htbp]
    \centering
    \resizebox{0.45\textwidth}{!}{
    \begin{tabular}{l|l|l|l}

       Training Set & Metric & Synth Test & Aria Test \\
    
      \hline
      \hline
    
                   & P50 $\downarrow$ & $8.45 \pm 0.89$ & $2.72 \pm 0.16$ \\
    
        Big Aria & P75 $\downarrow$ & $11.70 \pm 0.93$ & $4.10 \pm 0.23$ \\
    
                   & P95 $\downarrow$ & $19.05 \pm 1.00$ & $8.84 \pm 0.99$ \\
    
      \hline
                   & P50 $\downarrow$ & $11.79 \pm 2.50$ & $3.96 \pm 0.20$ \\
    
        Small Aria & P75 $\downarrow$ & $17.05 \pm 3.54$ & $5.93 \pm 0.29$ \\
    
                   & P95 $\downarrow$ & $27.43 \pm 4.72$ & $13.42 \pm 1.14$ \\
    
      \hline
                   & P50 $\downarrow$ & $2.99 \pm 0.10$ & $12.24 \pm 1.07$ \\
    
        Synth & P75 $\downarrow$ & $4.46 \pm 0.16$ & $18.35 \pm 1.28$ \\
    
                   & P95 $\downarrow$ & $8.06 \pm 0.51$ & $32.30 \pm 1.49$ \\

    \end{tabular}
    }
    \caption{\textbf{Baseline comparison.} Gaze estimation results on Synth and Aria test sets, without optical effects. All errors in degrees. }
    \label{tab:base_case_results}
\end{table}

\noindent\textbf{Direct comparison of errors}
First, we consider the ``same domain'' results, of real-trained models evaluated on real data, and synthetic-trained models evaluated on synthetic, from Tab. \ref{tab:base_case_results}.
We observe comparable performance between Synth and Big Aria, when tested on their respective test sets. 
Overall, the P50, P75 and P95 of both models are within standard deviations of each other. 
The discrepancies are larger with the P95 result, which could indicate that the synthetic training set still needs more identities and more diversity to include hard test cases. 
Errors on the Small Aria trained models (evaluated on Aria) are higher, which is expected for a model trained on fewer identities trying to generalize to a test set with many unseen identities.

There are many factors affecting the absolute performance numbers of a model, so these experiments primarily demonstrate that the synthetically-trained models are of similar magnitude to real-trained model error. 
Because there is a difference in scale of the errors, the remaining experiments  focus on showing that relative performance differences are preserved.

We briefly note the results on the ``different domain'' case where Aria-trained models are evaluated on Synth data and vice-versa.
These are uniformly higher than the same-domain case, demonstrating that a domain gap between the two datasets does exist.
Ideally synthetic data would be realistic enough to have zero domain gap with real data, but we do not require this to be predictive of relative performance differences.

\subsection{Effect of Optical Trends on Performance}
\label{ssec:novel-camera-experiments}

\begin{figure}[t]
    \centering
    \includegraphics[width=0.5\textwidth]{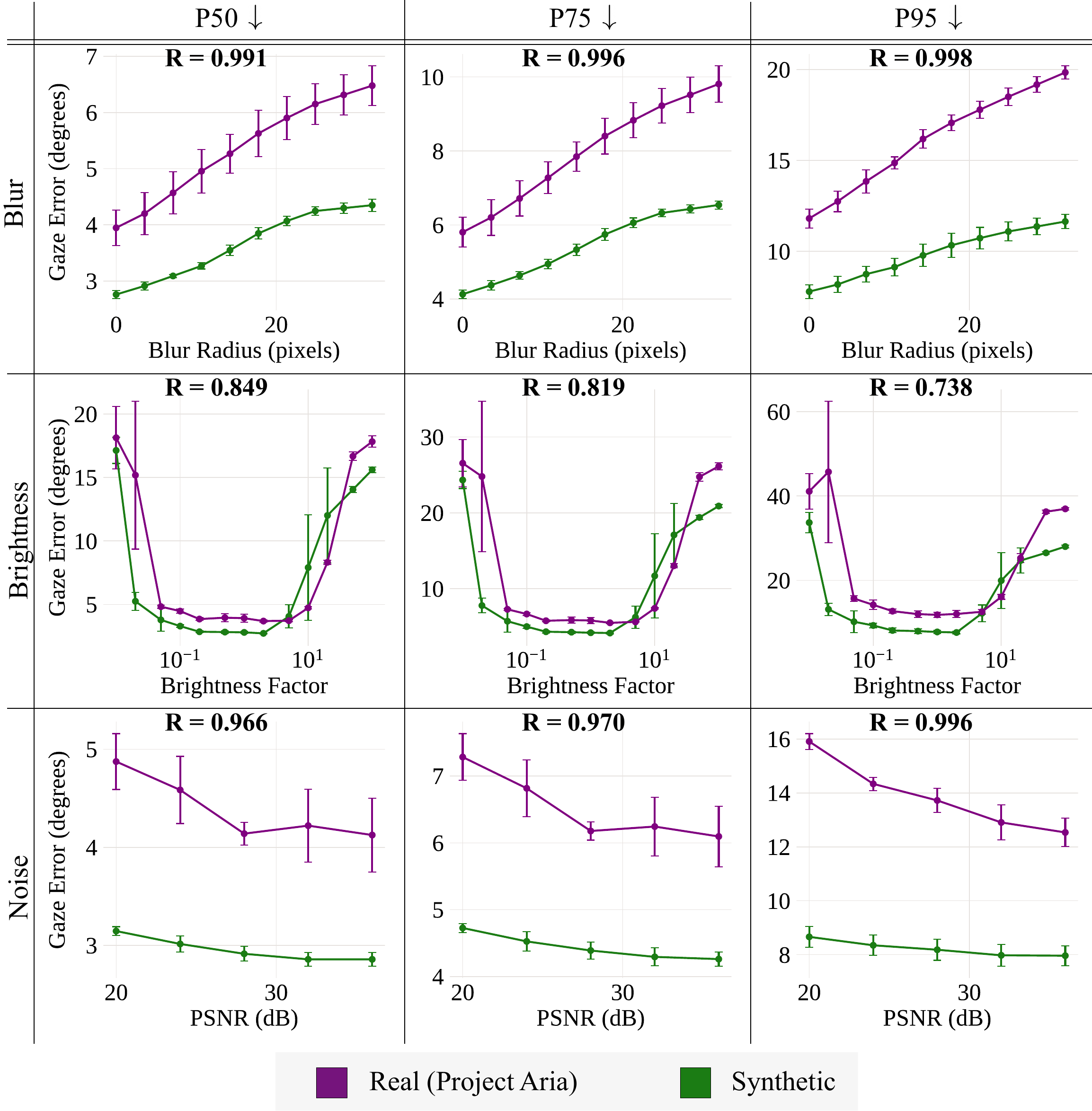}
    \caption{\textbf{Optical trends. } Model performance with respect to changing blur, brightness and noise are plotted, separately for each percentile metric. Correlation R-scores between real-trained and synthetic-trained models are given above each graph, showing high degree of correlation. }
    \label{fig:optical-trends}
\end{figure}

In this section, we investigate the extent to which our synthetic data generator can predict relative performance differences in model performance, as we vary optical effects like blur, brightness and noise.
The synthetic datasets add these effects as described in Sec.~\ref{ssec:optics}.
For the Aria datasets, all of these optical effects can be accomplished with 2D image manipulation, so we apply them identically to the synthetic images.
Due to the high compute time for training Big Aria models, in this section we only compare between Small Aria and Synth datasets.
Visualizations of sample input images are in the Supplement Sec. \ref{ssec:blur_viz}-\ref{ssec:noise_viz}.

\begin{figure*}[t]
\centering
\includegraphics[width=0.95\linewidth]{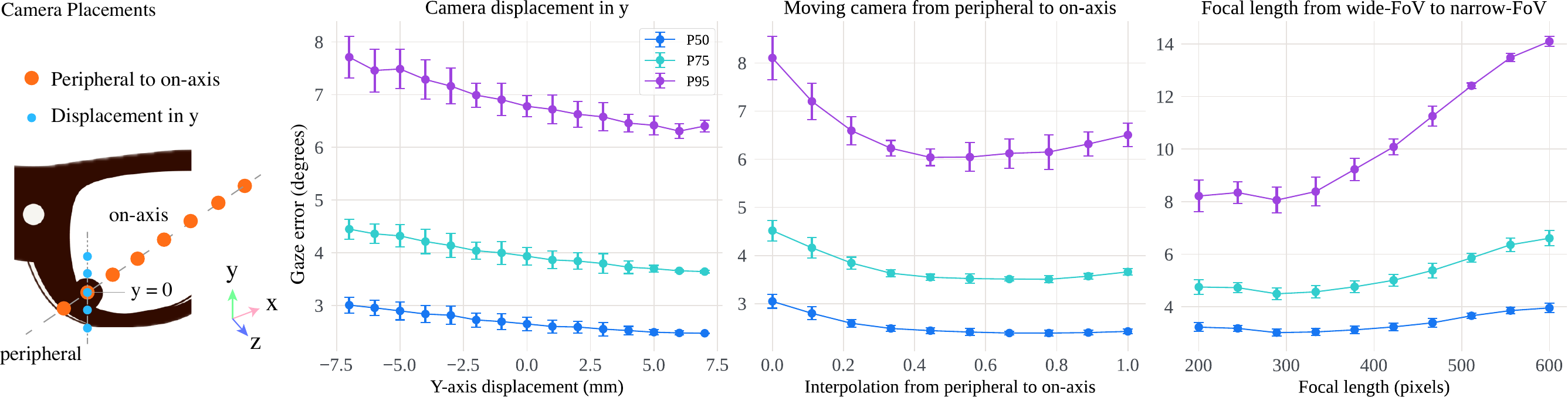}
\caption{\textbf{Novel camera experiments.} At left, schematic of moving cameras vertically (blue) and from peripheral to on-axis (orange). The first two graphs show model performance as a result of this changing camera location. The final graph shows results of model perfomance with changing focal length, but fixed camera location. No comparison with Aria data given, since all results require novel views.}
\label{fig:camera-viewpoint}
\end{figure*}

\paragraph{Blur}
In this experiment, we blur our training and test set pairs with increasing blur, from a range of 0 to 32 pixels. 
Fig.~\ref{fig:optical-trends} shows degrading performance for models trained on both Synth and Aria datasets as blur increases.
This is expected as the loss of high-frequency information provides less information for determining the exact gaze direction.
Qualitatively, the shape of the performance degradation in the graph is very similar between Aria-trained models and Synth-trained models, except for an overall scale factor.
Quantitatively, the correlation between synthetic and real trained results is above 0.99 R-score (Fig.~\ref{fig:optical-trends}).
With such a high correlation, a degradation of X\% for a given amount of blur in a synthetically trained model is very predictive of what we see in the real-trained model.
We additionally present cross-testing results in Supplement Sec. \ref{ssec:cross_testing}, where models are trained on one dataset (Aria or Synth) and tested on the other.

\paragraph{Brightness}
In this experiment, we consider the effect of modifying illumination brightness on eye tracking.
Brightness is modified in the range $[0.01, 100]$, which for 8-bit images ranges from very dark to almost completely saturated.
Fig.~\ref{fig:optical-trends} shows degrading performance for models trained on both Synth and Aria datasets at both extremas, when the image signal is weak and or saturated (respectively). 
We again see relatively high correlation between the synthetic and real results, with scores above 0.7.
We hypothesize that the correlations are negatively impacted by the extrema where performance ``falls off a cliff'', with median error above 15 degrees not being much better than random chance.
Qualitatively, the threshold-value of brightness change at which performance rapidly decreases is similar between real and synthetic results.

\paragraph{Noise}
In this experiment, we add increasing amounts of gaussian noise to our training and test set images. 
Fig.~\ref{fig:optical-trends} shows that eye tracking performance degrades as noise levels increase (PSNR decreases).
While the brightness experiments degrade SNR via saturation or quantization, adding gaussian noise leads to more predictable changes in performance, and we again see high correlation between real and synthetic data with R-score greater than 0.96.
In Supplement Sec. \ref{ssec:noise_brightness}, we present additional results that investigate performance trends for images corrupted by noise at different brightness levels.

\subsection{Novel viewpoint simulation}

The above experiments all have simulated data for which comparable real data exists, through 2D postprocessing of the Aria images to add optical effects.
In this section, we demonstrate the extrapolation of the simulator beyond situations where we can compare with real data.
Here, we rely on the 3D nature of the reconstructions, and the ability to do novel view synthesis from arbitrary input camera locations.
We perform 3 experiments, with results in Fig.~\ref{fig:camera-viewpoint}. 

In the first two experiments, we vary the camera location.
The first experiment moves the camera in the vertical direction, upwards and downwards from the nominal Aria camera location in steps of 1mm. 
The second experiment takes camera locations equally spaced along a straight line in 3D, starting from the nominal Aria camera location and ending at an on-axis camera directly in front of the eye.
The camera locations for these experiments are shown in Fig.~\ref{fig:camera-viewpoint}.

The Aria camera is positioned well off-axis, to the side (temporally) and downwards from a direct view of the eye.
ET is more difficult from this oblique viewpoint.
Both of these experiments confirm that as the camera is moved closer to on-axis (upwards in the first experiment, and towards the center in the second experiment) the performance improves.
Additionally, we see the performance saturate for the second experiment, indicating that an intermediate on axis view is as good or better than a directly on-axis view.

In the third experiment, we vary the focal length of the camera from 200 pixels to 600 pixels, (original Aria focal length is 270 pixels).
By increasing the focal length, we reduce the field of view (FoV) from one that is wider than the original Aria configuration to a significantly more zoomed-in view.
We see that performance improves slightly as we increase the resolution on the eye itself, up to a point where performance begins to degrade due to loss of FoV where the eye is not always visible.

In these experiments, we do not claim to find an optimal configuration for eye tracking.
The system designer must trade off performance of the system predicted by our simulator versus other constraints like system integration within the glasses frame.
For example, in the first two experiments, the camera locations are not physically located on the frame of the glasses, and in the third experiment, smaller field of view may be desirable for eye tracking, but prevents use of the cameras for other applications like face tracking.
Instead, we provide a tool whereby system designers can get signal as to what designs are good or bad, without having to do a complete build of a prototype to gather data.
\section{Discussion}
\label{label:discussion}

While this work focuses solely on data generation, high-quality, diverse, and representative data is fundamental to improving ML-based gaze estimation quality~\cite{wood2015rendering}, which in turn enables the development of faster, better networks~\cite{kim2019nvgaze}. We envision that our synthetic data generator can be leveraged to develop gaze estimators for novel camera designs.

 One advantage of our method based on NeRF is that we can render with flexible camera projections, including non-pinhole models and complex PSFs by sampling multiple rays per pixel.
This flexibility is crucial because, as our experiments confirm, optimizing optical metrics like the PSF blur in isolation does not directly enhance end-to-end performance in computer vision systems~\cite{chen2025isotropic,yang2023aberration,tseng2021differentiable}. 
Instead, it requires balancing with constraints such as form factor, weight, power consumption, and cost.

\noindent\textbf{Limitations} 
Since our framework is built upon real identities, it is subject to the population diversity captured, reducing its direct generalizability to real-world settings.
Furthermore, the system is constrained to gaze directions within the range captured from the participants of $\pm35^\circ$ due to the design of the light dome.  
It does not support extrapolation beyond this limit.
Our capture setup employs LEDs with a wavelength of 850 nm, and would require further work to simulate alternative wavelength conditions, such as narrow-band illumination or different spectral configurations.
This consideration is important because human skin exhibits differing reflections under infrared illumination~\cite{nair2020riteyes}, and our system is limited to capturing only at 850 nm.
Finally, our framework primarily addresses monocular scenarios by capturing only the right eye.
Extension to binocular eye tracking remains to be explored.

{
    \small
    \bibliographystyle{ieeenat_fullname}
    \bibliography{main}
}

\clearpage
\setcounter{page}{1}
\maketitlesupplementary

\section{Visualization of Experimental Inputs}
\label{ssec:experiment_viz}

In this section, we present example inputs to our experiments in Sec.~\ref{ssec:novel-camera-experiments} in the Main Paper. 

\subsection{Blur}
\label{ssec:blur_viz}

\begin{figure}[ht]
    \centering
    \includegraphics[width=0.48\textwidth]{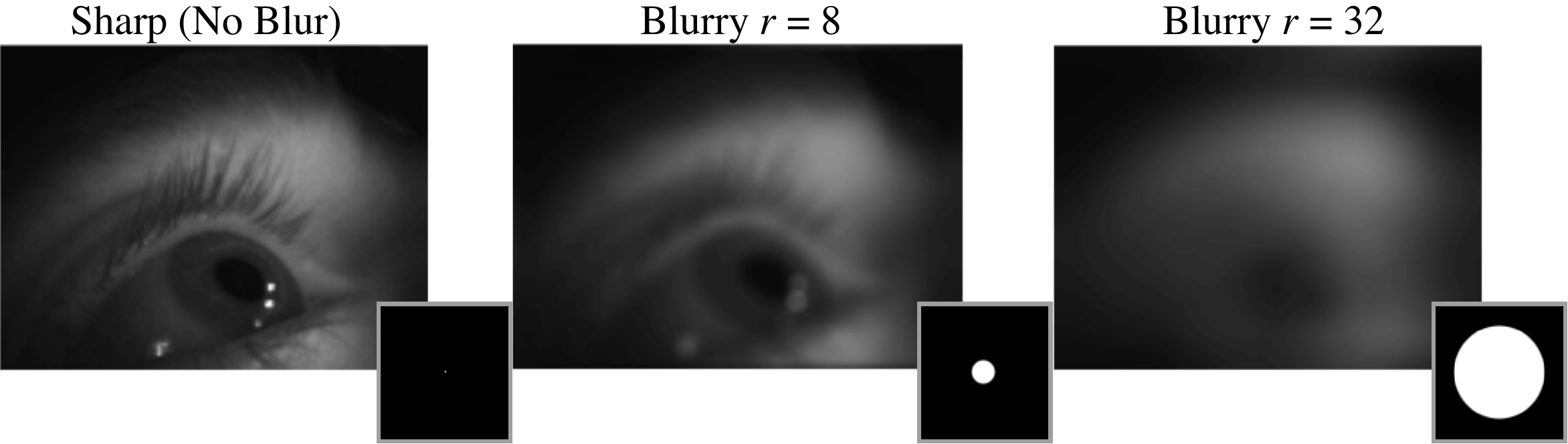}
    \caption{
    \textbf{Sample synthetic images with different blurriness.} The blur kernels are shown on the bottom right beside each image.}
    \label{fig:SM-blur-example}
\end{figure}

In this experiment, we blur our training and test set pairs with increasing aperture blur, from a range of radius $r \in [0, 32]$ pixels. 
The case $r=0$ indicates no blur, i.e., the original sharp image.
Fig. \ref{fig:SM-blur-example} presents example inputs with varying levels of blurriness. Results for the overall trend can be found in Fig. \ref{fig:optical-trends} of the Main Paper. 

\subsection{Brightness}
\label{ssec:brightness_viz}

\begin{figure}[ht]
    \centering
    \includegraphics[width=0.5\textwidth]{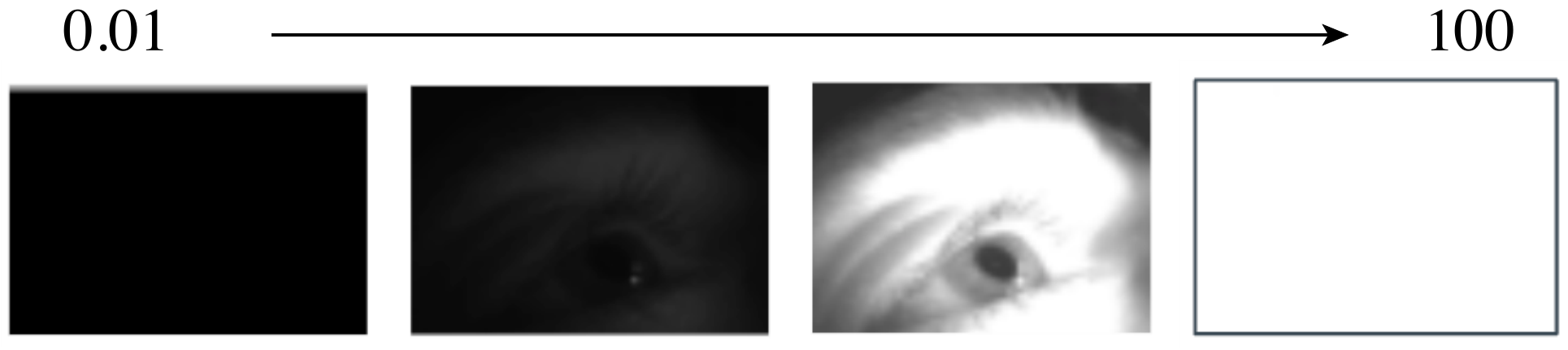}
    \caption{
    \textbf{Sample synthetic images with different brightness level} }
    \label{fig:SM-brightness-example}
\end{figure}

In this experiment, we consider the effect of modifying illumination brightness on eye tracking.
Brightness is modified by scaling input images with scalar factors in the range $[0.01, 100]$.
For 8-bit images, this operation causes the image intensities to range from very dark to almost completely saturated. 
We choose 8-bit quantization as that is the bit-depth used for the ET cameras on the Aria glasses. 
Fig. \ref{fig:SM-brightness-example} presents example inputs with varying levels of brightness, with special emphasis on the experiments nearopposite ends of the brightness extremas: very dark and very bright. 
We chose 0.01 to be the lower bound, as it gave an average intensity of < 1  over the brightness adjusted datasets. 
We chose 100 to be the upper bound, as it gave an average intensity of > 254 over the brightness adjusted datasets.
Figure \ref{fig:SM-brightness-example} shows that as brightness increases, the periocular region of the eye saturates first, while the iris and pupil saturate later, retaining details.  
This phenomenon may explain the central plateau region in gaze estimation performance as a function of brightness, as brightness changes are not yet big enough to discard key features like the pupil and iris.
Results for the overall trend can be found in Fig. \ref{fig:optical-trends} of the Main Paper. 

\subsection{Noise}
\label{ssec:noise_viz}

\begin{figure}[ht]
    \centering
    \includegraphics[width=0.5\textwidth]{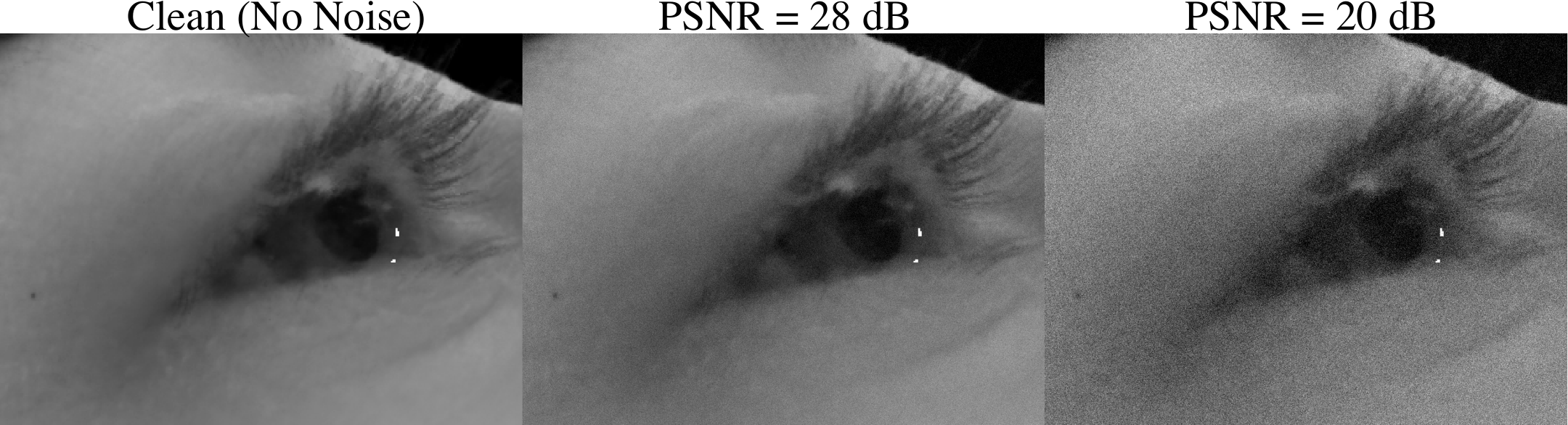}
    \caption{
    \textbf{Sample synthetic images with different noise level} }
    \label{fig:SM-noise-example}
\end{figure}

In this experiment, we add increasing amounts of gaussian noise to our training and test set images such that PSNR of resulting noisy images reaches a low bound of 20 dB. 
Figure \ref{fig:SM-noise-example} presents example inputs with varying levels of noise: a clean image, a moderately noisy image (28 dB), and the noisiest image (20 dB).
Results for the overall trend can be found in Fig. \ref{fig:optical-trends} of the Main Paper. 

\subsection{Focal Length}
In this experiment, we vary the focal length $f$ of the camera from 200 pixels to 600 pixels, (original Aria focal length is 270 pixels).
By increasing the focal length, we reduce the field of view (FoV) from one that is wider than the original Aria configuration to a significantly more zoomed-in view.
Fig. \ref{fig:SM-noise-example} presents example inputs with varying focal lengths. 
We observe that as the focal length increases, the FoV decreases, constraining our view of the eye.  
This corresponds to the increase in gaze error shown in Fig.~\ref{fig:camera-viewpoint} of the Main Paper, where performance begins to degrade due to a loss of FoV that causes the eye to be intermittently visible.

\label{ssec:FoV_viz}
\begin{figure}[ht]
    \centering
    \includegraphics[width=0.48\textwidth]{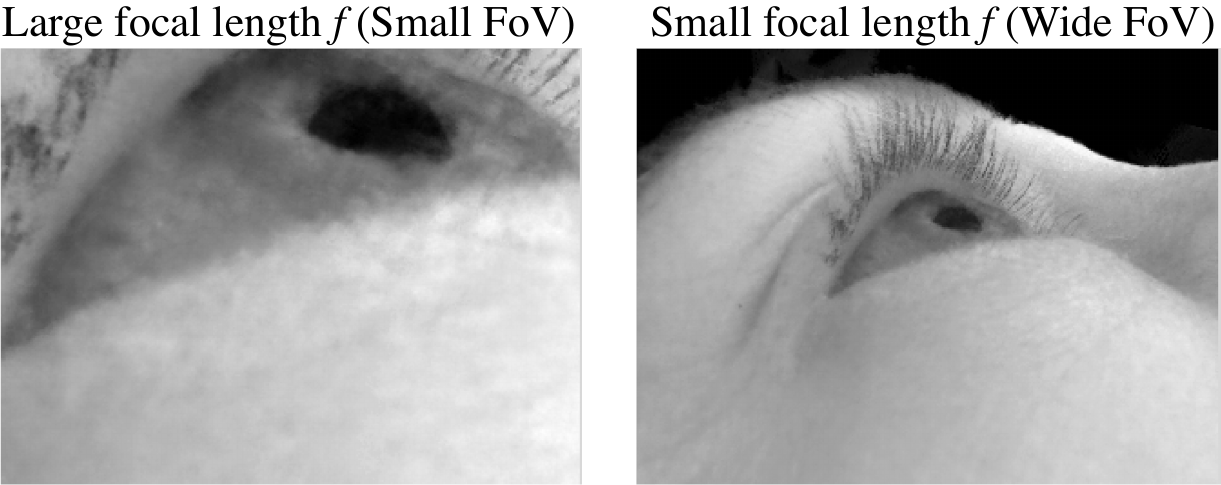}
    \caption{
    \textbf{Sample synthetic images with focal length 200 pixels (left) and 600 pixels (right).} }
    \label{fig:SM-fov-example}
\end{figure}

\section{Additional Experimental Results}
\label{sec:more_experiments}
\subsection{Cross Testing Results}
\label{ssec:cross_testing}
\begin{figure}
  \centering
  \includegraphics[width=0.5\textwidth]{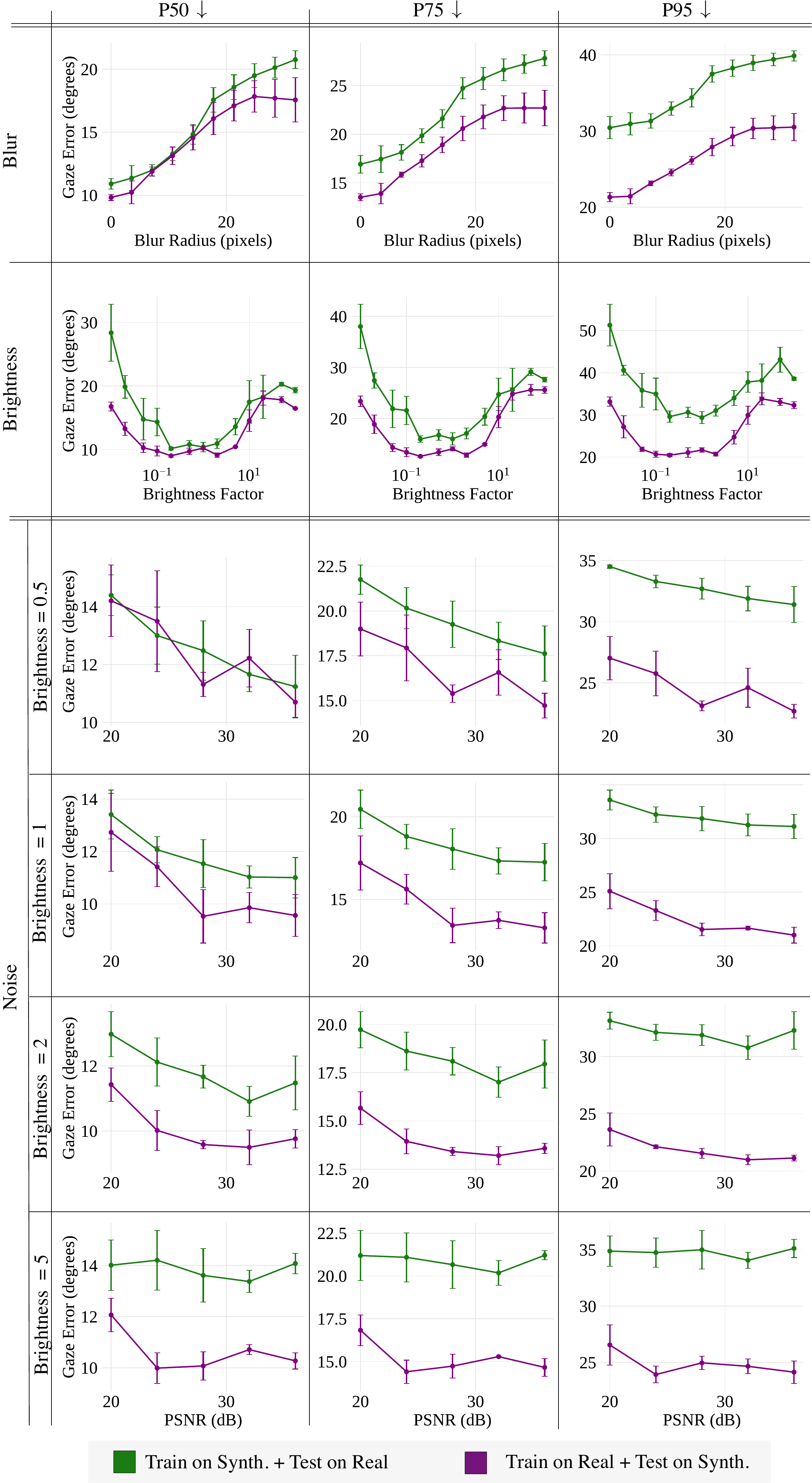}
  \caption{\textbf{Cross testing experiments of optical trends between synthetic and real data.} See Tab. \ref{tab:SM-optical-trends-correlation-all} for correlation scores.}
  \label{fig:SM-Optical-trends}
\end{figure}

We conducted cross-testing experiments to further investigate the correlation between synthetic and real data implications onto gaze estimation.
In this experiment, models are trained on one dataset (Aria or Synth) and tested on the other. 
The qualitative results are shown in Fig. \ref{fig:SM-Optical-trends}, and the quantitative correlation scores are summarized in Tab. \ref{tab:SM-optical-trends-correlation-all}. 
The results show that our framework is able to predict the performance change trend across population subject to the blur, brightness and noise change (when brightness is 1), as evidenced by high correlation score, except for the very-high-brightness (brightness factor = 5) scenario.

\begin{table}[h]
  \centering
  \resizebox{0.49\textwidth}{!}{
   \begin{tabular}{l|c|c|c}
         Experiment & P50 $\downarrow$  & P75 $\downarrow$  & P95 $\downarrow$\\
         \hline\hline
         Blur                       & 0.9673 & 0.9777 & 0.9778\\
         Brightness                 & 0.8224 & 0.8157 & 0.8021\\
         Noise (Brightness = 0.5)   & 0.8688 & 0.8940 & 0.8583\\
         Noise (Brightness = 1)     & 0.9481 & 0.9562 & 0.9613\\
         Noise (Brightness = 2)     & 0.9207 & 0.9252 & 0.8069\\
         Noise (Brightness = 5)     & 0.0135 & 0.1176 & 0.0719\\
  \end{tabular}
  }
  \caption{\textbf{Correlations (R) for cross testing experiments.} }
  \label{tab:SM-optical-trends-correlation-all}
\end{table}

\subsection{Noise Experiment as a Function of Brightness}
\label{ssec:noise_brightness}
Our noise experiments presented in Fig. \ref{fig:optical-trends} of the Main Paper use a single brightness configuration, although brightness and noise may have coupled effects.  
This is evidenced by the fact that PSNR metrics can be similar for two image pairs even when their average intensities differ.  
We chose to hold brightness constant in our noise experiments because solely adding Gaussian noise leads to more interpretable changes in performance.
In particular, we keep brightness = 1, which gives the unscaled original images from Project Aria, and direct renders from our framework.

Here, in this section, we present additional results that investigate performance trends for images corrupted by noise at different brightness levels.
We choose brightness levels of $0.5, 1, 2, 5$ corresponding to the performance plateau region in Fig. \ref{fig:optical-trends} of the Main Paper, since this range is neither as extreme as full saturation nor as low as no signal, making it a more feasible option for real-world scenarios.

\begin{figure}
  \centering
  \includegraphics[width=0.5\textwidth]{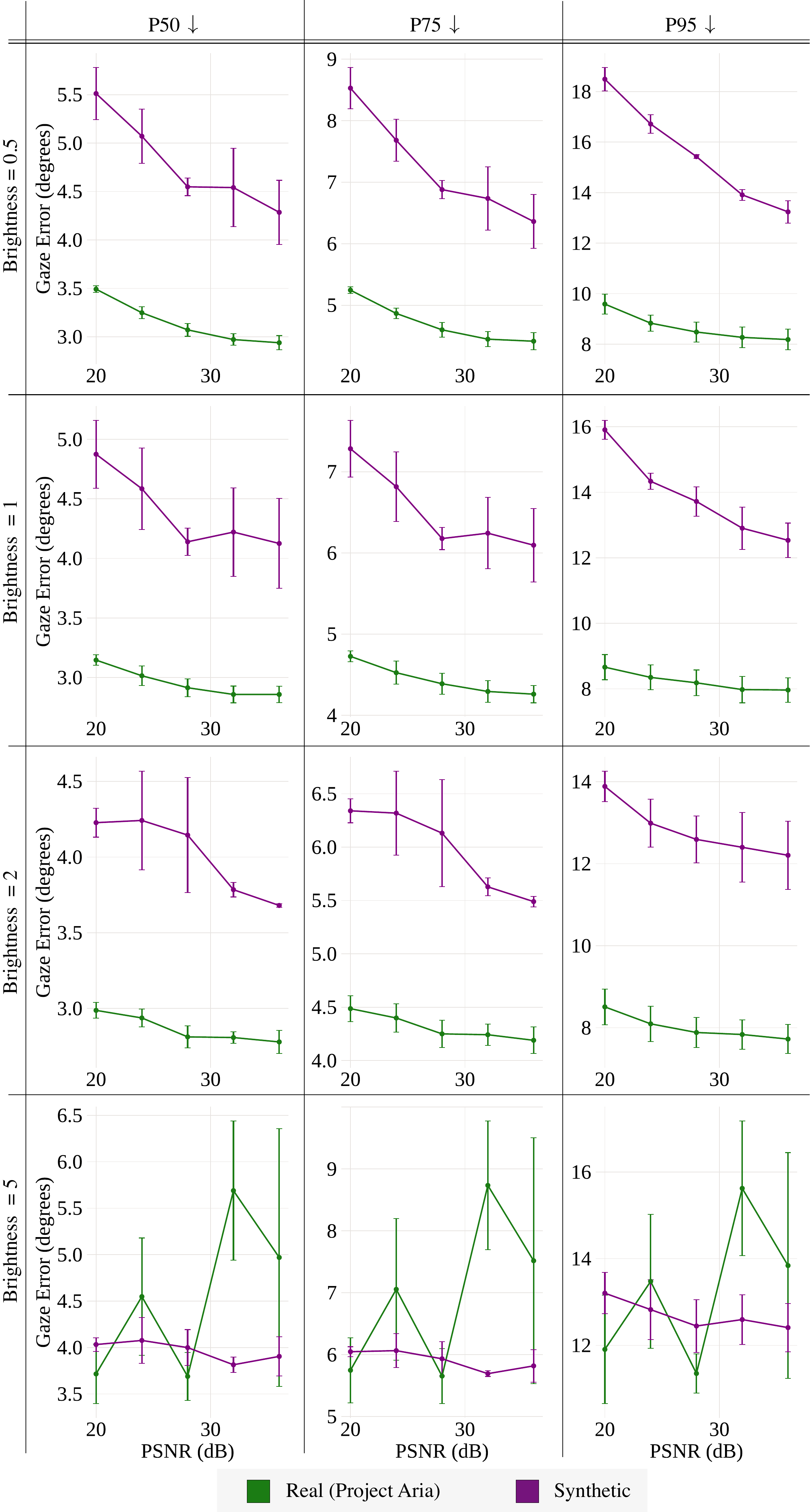}
  \caption{\textbf{Noise experiments as a function of brightness.} See Tab. \ref{tab:SM-noise-correlation-all} for correlation scores.}
  \label{fig:SM-noise-experiment}
\end{figure}

Qualitative results are shown in Fig. \ref{fig:SM-noise-experiment} , and quantitative correlation scores are summarized in Tab. \ref{tab:SM-noise-correlation-all}. 
We see that for relatively low-intensity cases (brightness = $0.5, 1$), our framework is able to capture noise trends. 
However, for the higher intensity cases (brightness = $2, 5$), the correlation decreases.

\begin{table} [h]
  \centering
  \resizebox{0.45\textwidth}{!}{
   \begin{tabular}{l|c|c|c}
         Experiment & P50  & P75  & P95 \\
         \hline\hline
         Brightness = 0.5   & 0.9845 & 0.9911 & 0.9683\\
         Brightness = 1     & 0.9656 & 0.9698 & 0.9957\\
         Brightness = 2     & 0.8004 & 0.8485 & 0.9989\\
         Brightness = 5     & -0.7864 & -0.7632 & -0.2656\\
  \end{tabular}
  }
  \caption{\textbf{Correlations (R) for noise experiment.} }
  \label{tab:SM-noise-correlation-all}
\end{table}

\clearpage

\clearpage

\end{document}